\documentclass{article}
\usepackage{spconf,amsmath,graphicx,hyperref}
\usepackage{amsfonts}
\makeatletter
\let\oldthebibliography\thebibliography
\def\thebibliography#1{%
  \oldthebibliography{#1}%
  \fontsize{9pt}{10pt}\selectfont    
  \setlength{\itemsep}{0pt plus 0.2pt}%
  \setlength{\parskip}{0pt}%
  \setlength{\parsep}{0pt}%
}
\makeatother
\usepackage{pgfplots}
\usepackage{booktabs}
\usepackage{graphicx}
\usepackage{adjustbox}  
\pgfplotsset{compat=1.18}
\usepgfplotslibrary{groupplots}
\usepackage{tikz}
\usepgfplotslibrary{statistics}   
\usepackage{CJKutf8}
\usetikzlibrary{arrows.meta,positioning}
\usepackage{multirow} 
\usepackage{pgf} 
\newcommand{\pgfres}[1]{%
  \pgfmathparse{#1}%
  \pgfmathprintnumber[fixed, precision=2]{\pgfmathresult}%
}

\newcommand{\smallheading}[1]{\vspace{0.1cm}\noindent\textbf{#1}\ \ }

\title{Bayesian Low-Rank Factorization for Robust Model Adaptation}
\name{Enes Yavuz Ugan$^{\star }$ \qquad Ngoc-Quan Pham$^{ \dagger}$ \qquad Alexander Waibel$^{\star\dagger}$}
\address{$^{\star}$Interactive Systems Lab, Karlsruhe Institut of Technology (KIT), Germany \\
      $^{\dagger}$InterACT, Carnegie Mellon University (CMU), USA}

\begin{document}
%
\maketitle
\begin{abstract}
Large speech foundation models achieve strong performance across many domains, but they often require adaptation to handle local needs such as code-switching, where speakers mix languages within the same utterance.
Direct fine-tuning of these models risks overfitting to the target domain and overwriting the broad capabilities of the base model. 
To address this challenge, we explore Bayesian factorized adapters for speech foundation models, which place priors near zero to achieve sparser adaptation matrices and thereby retain general performance while adapting to specific domains.

We apply our approach to the Whisper model and evaluate on different multilingual code-switching scenarios.
Our results show only minimal adaptation loss while significantly reducing catastrophic forgetting of the base model. 
Compared to LoRA, our method achieves a backward gain of 54\% with only a 4\% drop on the new domain.
These findings highlight the effectiveness of Bayesian adaptation for fine-tuning speech foundation models without sacrificing generalization.

\end{abstract}
\section{Introduction}
Large-scale multilingual speech foundation models \cite{radford2023robust, barrault2023seamless} have rapidly become the de facto backbones of modern automatic speech recognition (ASR) systems. Their success stems from (i) pre-training on tens of thousands of hours of heterogeneous audio, (ii) joint subword vocabularies enabling cross-lingual sharing, and (iii) transformer decoders capable of leveraging long-range linguistic context and are trained on thousands of transcripts. 
Yet even these billion-parameter models can exhibit brittleness when confronted with out-of-domain (OOD) data, such as low-resource languages or code-switching speech \cite{ ebrahimi2023findings, suhm1994towards, stuker2003multilingual}.
Adapting them to new domains while preserving their broad generalization ability remains a challenge, especially when access to the original pre-training data is restricted or not available at all.

Recently, Low-Rank (LoRA) adaptation \cite{pham2021efficient, hu2022lora} has emerged as a practical approach for domain adaptation, inserting small trainable matrices into frozen model weights enabling efficient fine-tuning with minimal memory overhead.
While LoRA has been promoted as a way to efficiently extend a model's capacity without catastrophic forgetting, this stability only holds when adapted weights are kept separate from the base model.
If the LoRA weights are merged into the model weights for inference (a common practice for deployment efficiency), even a few steps of fine-tuning on a distributionally distant dataset can cause severe degradation on the model's original capabilities \cite{ugan2025weight}.
The desiderata is to minimize this compromise, when the real world speech applications have to be especially flexible to be able to handle both the specialized domain as well as the general one~\cite{huber2023end}. 
For example, a particular use-case is to adapt the models for code-switching scenarios, but operating in other languages is still necessary. This speech contains a lot of sociolinguistic importance and conveys meaningful information \cite{dougruoz2023survey, myers1977bilingual} that can be used further in other tasks.
Simply fine-tuning may improve performance on the target language mix, but at the cost of noticeably worse recognition on other language pairs or monolingual speech.

In this work, we explore an alternative: \textit{Bayesian Low-Rank Adaptation (BLoRA)} for speech foundation models.
By placing a zero-mean, small-variance Gaussian prior over LoRA parameters, we regularize the adaptation toward "sparser" weight matrices, thereby reducing overfitting to the new domain and limiting destructive changes to the base model's latent space.
We show that, in \textit{single-step adaptation}, BLoRA can better retain base model accuracy after domain adaptation compared to standard LoRA.
Our contributions are:
\begin{itemize}\setlength\itemsep{0.1em}
    \item The first application of Bayesian Low-Rank Adaptation to speech foundation models, enabling prior-based regularization of domain adaptation.
    \item Empirical evidence that BLoRA improves retention of base model performance in single-step fine-tuning on diverse code-switching ASR tasks.
    \item Analysis of resulting weight matrices in terms of sparsity and distribution.
\end{itemize}

\section{Related Work}
Earlier studies on multilingual ASR investigated cross-lingual acoustic modeling and articulatory feature sharing across languages \cite{schultz1996lvcsr,schultz2001experiments,stuker2003integrating}, laying the foundation for today's code-switching and low-ressource adaptation research.

\smallheading{Code-Switching}
There is rising interest in code-switching ASR as can be seen in \cite{hamed2020arzen, abdallah2024leveraging} in which the authors focus on Egyptian Arabic and Tunisian Arabic code-switching with English. 
Authors in \cite{lyu2010seame} work with Mandarin-English language pairs, while \cite{weller2022end} focus on Spanish-English and German-English in \cite{ugan2024decm}.
Dedicated approaches at solving code-switching speech recognition recently cover different approaches such as text based ones \cite{hussein2023textual, hamed2022investigating}.
Both explore lexical swaps into monolingual corpora, randomly, or driven by theory.
In \cite{du2021data} data augmentation is extended into the audio domain.
Authors splice existing code-switched utterances and synthesize speech after world-level translations, or insertions, thus improving both acoustic and language models, however, their approach depends on the availability of parallel translation data and good text-to-speech (TTS) synthesis.
Authors in \cite{ugan2022language} showed that simple inter-sentential concatenation of audio and text pairs boosts language-agnostic ASR performance on both mixed language and monolingual tests.
A speech-chain loop in which ASR and TTS models improve each other iteratively, was introduced in \cite{nakayama2021code}, yielding improvements for code-switching WER.
In most of the mentioned work authors do not investigate model performance on some OOD monolingual data present in the code-switching scenarios and none of the works consider third party languages not present in the adaptation scenarios.


\smallheading{Domain Adaptation for Speech Foundation Models}
Adapting large-scale speech foundation models to new domains without degrading their general-purpose performance is an active area of research. 
Parameter-efficient tuning methods such as Low-Rank Adaptation (LoRA) \cite{hu2022lora, pham2021efficient} and related adapter-based approaches \cite{pfeiffer2020adapterfusion} insert small trainable modules into frozen weight tensors, reducing memory and compute overhead.
Several works \cite{pham2022adaptive, rolland2024shared,liu2023kit} have applied LoRA-style adapters for ASR and speech translation, showing strong gains on the target domain. 
However, when LoRA weights are merged back into the base model for deployment, performance on the original training distribution often degrades sharply \cite{ugan2025weight}. 
This phenomenon is particularly problematic in scenarios where access to pre-training data is limited or restricted.

To mitigate forgetting, prior work has explored regularization based adaptation such as L2-SP \cite{xuhong2018explicit}, Elastic Weight Consolidation (EWC) \cite{kirkpatrick2017overcoming}, and Synaptic Intelligence (SI) \cite{zenke2017continual}, as well as multi-head or task-specific adapter routing \cite{pfeiffer2020adapterfusion, poth2023adapters}. 
While effective in some settings, these approaches often require either multiple forward passes or explicit storage of previous task data, making them less suitable for one-off adaptation in privacy- or license-restricted environments.
Our work targets this one-step adaptation regime, introducing a Bayesian formulation of LoRA that constrains parameter drift while still improving performance on the new domain.

\smallheading{Bayesian Fine-Tuning}
Bayesian neural networks and variational inference have been explored for improving model robustness and uncertainty estimation \cite{fortunato2017bayesian,blundell2015weight,wang2024blob,yang2023bayesian}, as well as in speech recognition \cite{xie2021bayesian}.
However, these techniques have not been applied to speech foundation models or LoRA-based tuning. 
Our approach is the first to apply a Bayesian prior to LoRA parameters in ASR, enabling prior-based regularization that improves retention of base model performance in a domain adaptation setting.

\section{Methodology}
\subsection{Low-Rank Adaptation}
\label{subsec:lora}
Whisper is a large-scale multilingual speech foundation model, consisting of a Transformer architecture and shared subword vocabulary across languages \cite{radford2023robust}.
Given the large parameter count of modern foundation models, domain adaptation is often performed via parameter efficient fine-tuning, reducing training cost and memory without increasing inference complexity.
In this work, we employ Low-Rank Adaptation (LoRA), which re-parameterizes a frozen weight matrix $W \in \mathbb{R}^{d_o \times d_i}$ as:
\[
W = W_0 + \frac{\alpha}{r} A B,
\]
with $A \in \mathbb{R}^{d_o \times r}$ and $B \in \mathbb{R}^{r \times d_i}$ are trainable low-rank matrices, $r \ll \min(d_i, d_o)$ is the rank, and $\alpha$ scales the update magnitude.  
This approach drastically reduces the number of trainable parameters while achieving strong downstream performance.

\subsection{Variational Fine-Tuning Objective}
\label{subsec:vlora}
When adapting a frozen model using a small set of trainable parameters $\theta$ (e.g., adapter weights), we treat $\theta$ as a set of latent variables and approximate their Bayesian posterior $q_\phi(\theta)$ via variational inference.
Maximizing the evidence lower bound (ELBO) yields:
\begin{equation}
\mathcal{L}_{\mathrm{ELBO}} =
\underbrace{\mathrm{CE}(y, \hat{y})}_{\text{data fit}}
+ \beta\, \underbrace{D_{\mathrm{KL}}\!\big[q_{\phi}(\theta) \,\|\, p(\theta)\big]}_{\text{complexity penalty}},
\end{equation}
where $p(\theta)$ is a chosen prior and $\beta \ge 0$ balances the fidelity to the data against the closeness to the prior.  
$\mathrm{CE}$ denotes the token-level cross-entropy loss, and $D_{\mathrm{KL}}$ the Kullback–Leibler divergence.  
Setting $\beta=0$ yields a purely variational model, while $\beta>0$ imposes a Bayesian regularization toward the prior.

\subsection{Bayesian LoRA}
\label{subsec:blora}
We instantiate the above variational objective for LoRA adapters in Whisper.
In all experiments, we apply LoRA updates $\Delta W = AB$ to the query and key projection layers, using a rank $r=32$.
Unlike standard LoRA, we treat each element of $A$ and $B$ as a latent variable with a fully factorized Gaussian posterior:
\[
q_\phi(A_{ij}) = \mathcal{N}(\mu_{ij}, \sigma_{ij}^2), \quad
q_\phi(B_{ij}) = \mathcal{N}(\mu'_{ij}, {\sigma'_{ij}}^{2}).
\]
The learnable parameters $\phi = \{\mu, \log\sigma, \mu', \log\sigma'\}$ are optimized via the ELBO in Section~\ref{subsec:vlora}.  
Gradients are estimated using the reparameterization trick \cite{kingma2013auto}:
\[
A_{ij} = \mu_{ij} + \sigma_{ij} \,\epsilon_{ij}, \quad \epsilon_{ij} \sim \mathcal{N}(0,1),
\]
and analogously for $B_{ij}$.  

We adopt an isotropic Gaussian prior with $\sigma_p = 0.01$ ($\sigma^2=1e^{-4}$), and $\mu=0$. 
The mean values for $B$ are zero initialized and $\log\sigma$ is set to $-50$ yielding almost zero variance, effectively starting from the pre-trained base model \cite{hu2022lora}.  
The mean values for $A$ use Kaiming-uniform initialization, with $\log\sigma$ drawn uniformly from $[0, -4.5)$.
The resulting Bayesian-LoRA (BLoRA) loss is:
\begin{equation}
\begin{aligned}
\mathcal{L}_{\mathrm{BLoRA}}
&= \mathrm{CE}(y, \hat{y}) \\
&\quad + \beta \sum_{i,j} D_{\mathrm{KL}}\big[q_{\phi}(A_{ij}) \,\|\, p(A_{ij})\big] \\
&\quad + \beta \sum_{i,j} D_{\mathrm{KL}}\big[q_{\phi}(B_{ij}) \,\|\, p(B_{ij})\big],
\end{aligned}
\label{eq:blora-loss}
\end{equation}
For diagonal Gaussians the KL admits a closed form:
\[
D_{\mathrm{KL}} =
\frac12 \left(
\frac{\sigma^{2}}{\sigma_p^{2}} +
\frac{\mu^{2}}{\sigma_p^{2}} - 1 +
2\log\frac{\sigma_p}{\sigma}
\right),
\]
which we additionally normalize by the number of weights.

\subsection{Predictive Distribution}
A deterministic LoRA-adapted model produces:
\[
p_{\mathrm{det}}(y \mid x) =
p\!\big(y \mid x,\, W_0 + AB\big),
\]
where $W_0$ is the frozen base weight and $A,B$ are fixed point estimates.

In BLoRA, $(A,B)$ are random variables with posterior $q_{\phi}(A,B)$, yielding the Bayesian predictive distribution:
\[
p_{\mathrm{B}}(y \mid x) =
\mathbb{E}_{(A,B) \sim q_{\phi}} \big[
p\!\big(y \mid x,\, W_0 + AB\big)
\big].
\]
At inference, this can be approximated either by Monte Carlo sampling of $(A,B)$ or by using the posterior mean.  
Even the mean estimate can reduce overconfidence by incorporating the uncertainty learned during training.

\section{Experiments}
\subsection{Experiment Setup}
Although Whisper model has a very strong multilingual transcription performance it still is very limited in the accuracy when predicting code-switching speech.
This being the reason we choose Whisper’s \emph{large-v3-turbo} variant, which is considered a strong foundation for multilingual speech recognition.
Proving the general applicability of the approach we utilize three very different code-switching datasets in our experiments. 
Fisher \cite{weller2022end} is a Spanish-English dataset consisting of telephone conversations.
ArzEn \cite{hamed2020arzen} is a conversational dataset with Egyptian Arabic and English code-switching, and SEAME \cite{lyu2010seame} is a Mandarin-English corpora collected in South East Asia.
For our Backward evaluation we used the same datasets as described in \cite{ugan2025weight}.
The languages used for backward evaluation are English, German, Arabic, Turkish, Mandarin and Spanish.
For better readability we report the averaged scores for each model.

In all our experiments we apply LoRA with rank 32, applied on top of the query and key projection layers of the model.
For BLoRA we define the prior with $\sigma=0.01$ and $\mu=0$, thus encouraging sparser adapter weights.
For inference, we simply take the learned $\mu$ as the weights and do not sample the weights, effectively reducing the inference time and the number of weights.

We set the learning rate to 0.001 with 2000 warm-up steps. 
We train models for 30000 steps and choose the best model based on the validation set for our evaluations.

We evaluate our models on standard metrics such as Word-Error-Rate (WER), Character-Error-Rate (CER) and Mixed-Error-Rate (MER) depending on the dataset, without code-switching–specific metrics (e.g., PIER\cite{ugan2025pier}) since our approach targets general robust adaptation rather than code-switching specifically.
Our Baseline comparison is the LoRA adapted model which is applied as a standard domain adaptation technique on current foundation models.

\subsection{Fine-tuning considering Forgetting}
As the KL loss yields high values and we did not want it to dominate the overall training loss, we chose a moderate $\beta$ of $0.5$ (in Equation~\ref{eq:blora-loss}).
Table~\ref{tab:single_step} reports result for the off the shelf Whisper model (Base), fine-tuning with LoRA adapters (LoRA), and our proposed Bayesian-LoRA adapters (BLoRA).
To ensure robustness, we conduct experiments on three aforementioned datasets, training on the domain-specific splits, and reporting results on their respective test sets (In-domain), as well as on the average of the backward test sets.
\begin{table}[t]
\centering
\small
\setlength\tabcolsep{3pt}
\renewcommand{\arraystretch}{1.05}
\begin{tabular}{llccc}
\toprule
Domain & Method & In-domain & Backward & $\Delta$ \\
\midrule
\multirow{5}{*}{ArzEn} 
 & Base & 52.8  & \pgfres{(9.8+6+6+23.5+16.7+9.3+6.1)/7} & -- \\
 & LoRA & 34.65 & \pgfres{(15.38+5.8+10.11+49+43.5+83.61+29.04)/7} & +\pgfres{(33.78-11.06)} \\
 & BLoRA & 38.22 & 20.42 & +\pgfres{(20.42-11.06)} \\
\midrule
\multirow{5}{*}{SEAME} 
 & Base &  29.4 & \pgfres{(9.8+6+6+23.5+16.7+9.3+6.1)/7} & -- \\
 & LoRA & \pgfres{(14.84+20.66)/2} & \pgfres{(36.61+7.89+23.38+132.41+151.78+34.08+53.43)/7} & +\pgfres{(62.8-11.06)} \\
 & BLoRA &    \pgfres{(17.62+24.77)/2}  & \pgfres{(12.76+4.29+6.18+23.13+14.57+10.09+7.32)/7} & +\pgfres{(11.19-11.06)} \\
\midrule
\multirow{5}{*}{Fisher} 
 & Base & 29.4 & \pgfres{(9.8+6+6+23.5+16.7+9.3+6.1)/7} & -- \\
 & LoRA & 19.92 & \pgfres{(18.6+5.66+14.06+49.55+30.5+35.28+9.51)/7} & +\pgfres{(23.31-11.06)} \\
 & BLoRA & 20.73  & \pgfres{(12.27+5.2+6.03+20.15+15.4+8.62+6.12)/7} & \pgfres{(10.54-11.06)} \\
\bottomrule
\end{tabular}
\caption{Single-step domain adaptation results (WER/MER\% on adaptation set, WER\% on Backward sets, $\Delta$WER/CER). Lower is better.}
\label{tab:single_step}
\end{table}
Relative to the baseline Whisper model, both LoRA variants improve in-domain adaptation, with BLoRA offering substantially better retention.
The KL divergence loss in BLoRA acts as a regularizer by constraining the adapter weights to remain close to the prior.
Consequently, large weight updates are only encouraged when strongly supported by the adaptation distribution.
This behavior embodies the stability-plasticity tradeoff:
LoRA being highly plastic, quickly adapts but is prone to overfitting and forgetting, whereas BLoRA balances plasticity with stability, yielding comparable adaptation performance and far stronger preservation of the base model's generalization capabilities.
This tradeoff is most evident on SEAME data, where BLoRA retains 28\% relative improvement over the off-the-shelf baseline and reduces backward degradation from 62.8\% to nearly zero (0.13\%).
The most challenging case occurs on the Egyptian-English ArzEn dataset, where BLoRA reaches 20.42pp backward error, still outperforming LoRA by $\sim$40\%.

\subsection{Analysis of learned Weights}
To quantify the effects of our Bayesian prior, we analyze the sparsity of the learned $\Delta W$ matrices. 
Table~\ref{tab:sparsity} reports our four complementary metrics (i) \textit{Thresh}: the fraction of weights below $10^{-3}$ (absolute sparsity), (ii) \textit{Adaptive}: adaptive sparsity relative to the baseline's (LoRA) median scale (robust to trivial sclaing), (iii) \textit{Top-1}: energy concentration in the largest 1\% of weights (compression), and (iv) \textit{Hoyer}: Hoyers's index (scale invariant sparsity).
As there are many adapter matrices, for each layer $l$ we calculate the metric and average them:
\[
\;\overline{m} = \frac{1}{L} \sum_{l=1}^L f(\Delta W_l)\;,
\]
with $f$ depicting the metric which is calculated.
Compared to LoRA, BLoRa exhibits dramatically higher sparsity, with 99.7\% of the weights having values smaller than $1e^{-3}$, compared to only 4.1\%.

To account for simple scaling differences, leading to wrong conclusion in terms of sparsity, we calculate the adaptive sparsity.
Adaptive sparsity measures the fraction of weights below a threshold that is scaled to the baseline's (LoRA) own median magnitude, i.e., the cutoff adapts to each layer's scale. 
This makes the measure robust to trivial rescaling and highlights genuine shrinkage in the weight distribution.
For LoRA, the adaptive sparsity at $\tau=0.5$ is 0.26, consistent with a dense and broadly distributed set of learned weights.
In contrast, BLoRA reaches 0.999, meaning virtually all of its updates are smaller than half the baseline's median magnitude.
This indicates that BLoRA produces genuinely sparse updates, rather than a trivial rescaling of the LoRA distribution.
Even if we consider stronger scaling by factors of 0.25 instead of 0.5, BLoRA consistently saturates near ~1.0, confirming a re-shaped distribution with most weights effectively collapsed to zero.
\begin{table}[t]
\centering
\resizebox{\columnwidth}{!}{
\begin{tabular}{lcccc}
\toprule
Adapter & \textbf{Thresh@1e-3} & \textbf{Adaptive@0.5} & \textbf{Top-1\%E} & \textbf{Hoyer} \\
\midrule
LoRA    & 4.1\% & 0.26 & 9.2\%  & 0.22 \\
BLoRA   & 99.7\% & 0.999 & 37.5\% & 0.45 \\
\bottomrule
\end{tabular}
}
\caption{Sparsity analysis of LoRA vs Bayesian LoRA (BLoRA). Metrics averaged across all $\Delta W$ matrices.}
\label{tab:sparsity}
\end{table}

These findings are futher supported by the energy concentration: in LoRA, the top 1\% of weights account for only 9.2\% of the squared-norm energy, whereas in BLoRA they capture 37.5\%.
This indicates that BLoRA concentrates the effective signal in a small subset of parameters, leaving the majority near zero.

We also report Hoyer's sparsity in column four, which ranges from 0 (uniform dense distribution) to 1 (maximally sparse, only one none zero).
LoRA yields 0.22, consistent with a dense distribution, while BLoRA doubles this score to 0.45, confirming a concentration of energy in few parameters.

\section{Conclusion}
In this paper, we introduce the first (to the best of our knowledge) Bayesian LoRA in speech foundation models.
This enables researchers to effectively adapt a massive foundation model with only a small number of parameters and introduces prior knowledge to training.
This prior knowledge can be used in many ways.
In this work, we utilize it to learn sparse adapter matrices, these inject a noisy distribution epsilon, similar to weight drop, thus mitigation catastrophic forgetting while still achieving competitive results for fine-tuning on new distributions.
\section{Acknowledgment}
This work was supported in part by BMBF (01EF1803B - RELATER, HoreKa), the EU Horizon program (101135798 – Meetween; 101213369 – DVPS), and grants from Zoom VC (2nd author) and Interactive-AI.

\bibliographystyle{IEEEbib}
\bibliography{strings,refs}

\begin{thebibliography}{10}

\bibitem{radford2023robust}
Alec Radford, Jong~Wook Kim, Tao Xu, Greg Brockman, Christine McLeavey, and Ilya Sutskever,
\newblock ``Robust speech recognition via large-scale weak supervision,''
\newblock in {\em ICML}.

\bibitem{barrault2023seamless}
Lo{\"\i}c Barrault, Yu-An Chung, Mariano~Coria Meglioli, et~al.,
\newblock ``Seamless: Multilingual expressive and streaming speech translation,''
\newblock {\em arXiv preprint arXiv:2312.05187}, 2023.

\bibitem{ebrahimi2023findings}
Abteen Ebrahimi, Manuel Mager, Adam Wiemerslage, Pavel Denisov, Arturo Oncevay, Danni Liu, et~al.,
\newblock ``Findings of the second americasnlp competition on speech-to-text translation,''
\newblock in {\em NeurIPS 2022 Competition Track}. PMLR, 2023.

\bibitem{suhm1994towards}
Bernhard Suhm,
\newblock ``Towards better language models for spontaneous speech,''
\newblock in {\em Proc. ICSLP'94}, 1994.

\bibitem{stuker2003multilingual}
Sebastian Stuker, Tanja Schultz, Florian Metze, and Alex Waibel,
\newblock ``Multilingual articulatory features,''
\newblock in {\em 2003 IEEE International Conference on Acoustics, Speech, and Signal Processing, 2003. Proceedings.(ICASSP'03).} IEEE.

\bibitem{pham2021efficient}
Ngoc-Quan Pham, Tuan-Nam Nguyen, Sebastian St{\"u}ker, and Alexander Waibel,
\newblock ``Efficient weight factorization for multilingual speech recognition,''
\newblock {\em arXiv preprint arXiv:2105.03010}.

\bibitem{hu2022lora}
Edward~J Hu, Yelong Shen, Phillip Wallis, Zeyuan Allen-Zhu, Yuanzhi Li, Shean Wang, Lu~Wang, Weizhu Chen, et~al.,
\newblock ``Lora: Low-rank adaptation of large language models.,''
\newblock {\em ICLR}.

\bibitem{ugan2025weight}
Enes~Yavuz Ugan, Ngoc-Quan Pham, and Alexander Waibel,
\newblock ``Weight factorization and centralization for continual learning in speech recognition,''
\newblock {\em arXiv preprint arXiv:2506.16574}, 2025.

\bibitem{huber2023end}
Christian Huber, Tu~Anh Dinh, Carlos Mullov, Ngoc~Quan Pham, Thai~Binh Nguyen, Fabian Retkowski, Stefan Constantin, Enes~Yavuz Ugan, Danni Liu, Zhaolin Li, et~al.,
\newblock ``End-to-end evaluation for low-latency simultaneous speech translation,''
\newblock {\em arXiv preprint arXiv:2308.03415}, 2023.

\bibitem{dougruoz2023survey}
A~Seza Do{\u{g}}ru{\"o}z, Sunayana Sitaram, Barbara~E Bullock, and Almeida~Jacqueline Toribio,
\newblock ``A survey of code-switching: Linguistic and social perspectives for language technologies,''
\newblock {\em arXiv preprint arXiv:2301.01967}, 2023.

\bibitem{myers1977bilingual}
Carol Myers~Scotton and William Ury,
\newblock ``Bilingual strategies: The social functions of code-switching,''
\newblock 1977.

\bibitem{schultz1996lvcsr}
Tanja Schultz, Ivica Rogina, and Alex Waibel,
\newblock ``Lvcsr-based language identification,''
\newblock in {\em 1996 IEEE International Conference on Acoustics, Speech, and Signal Processing Conference Proceedings}. IEEE, 1996, vol.~2, pp. 781--784.

\bibitem{schultz2001experiments}
Tanja Schultz and Alex Waibel,
\newblock ``Experiments on cross-language acoustic modeling.,''
\newblock in {\em INTERSPEECH}, 2001.

\bibitem{stuker2003integrating}
Sebastian St{\"u}ker, Florian Metze, Tanja Schultz, and Alex Waibel,
\newblock ``Integrating multilingual articulatory features into speech recognition.,''
\newblock in {\em INTERSPEECH}, 2003.

\bibitem{hamed2020arzen}
Injy Hamed, Ngoc~Thang Vu, and Slim Abdennadher,
\newblock ``Arzen: A speech corpus for code-switched egyptian arabic-english,''
\newblock in {\em Proceedings of the twelfth language resources and evaluation conference}, 2020.

\bibitem{abdallah2024leveraging}
Ahmed Amine~Ben Abdallah, Ata Kabboudi, Amir Kanoun, and Salah Zaiem,
\newblock ``Leveraging data collection and unsupervised learning for code-switched tunisian arabic automatic speech recognition,''
\newblock in {\em (ICASSP)}, 2024.

\bibitem{lyu2010seame}
Dau-Cheng Lyu, Tien~Ping Tan, Engsiong Chng, and Haizhou Li,
\newblock ``Seame: a mandarin-english code-switching speech corpus in south-east asia.,''
\newblock in {\em Interspeech}, 2010.

\bibitem{weller2022end}
Orion Weller, Matthias Sperber, et~al.,
\newblock ``End-to-end speech translation for code switched speech,''
\newblock {\em arXiv preprint arXiv:2204.05076}, 2022.

\bibitem{ugan2024decm}
Enes~Yavuz Ugan, Ngoc-Quan Pham, and Alex Waibel,
\newblock ``Decm: Evaluating bilingual asr performance on a code-switching/mixing benchmark,''
\newblock in {\em Proceedings Of The 2024 Joint International Conference On Computational Linguistics, Language Resources And Evaluation (LREC-COLING 2024)}, 2024, pp. 4468--4475.

\bibitem{hussein2023textual}
Amir Hussein, Shammur~Absar Chowdhury, Ahmed Abdelali, Najim Dehak, Ahmed Ali, and Sanjeev Khudanpur,
\newblock ``Textual data augmentation for arabic-english code-switching speech recognition,''
\newblock in {\em 2022 IEEE Spoken Language Technology Workshop (SLT)}.

\bibitem{hamed2022investigating}
Injy Hamed, Nizar Habash, Slim Abdennadher, and Ngoc~Thang Vu,
\newblock ``Investigating lexical replacements for arabic-english code-switched data augmentation,''
\newblock {\em arXiv preprint arXiv:2205.12649}, 2022.

\bibitem{du2021data}
Chenpeng Du, Hao Li, Yizhou Lu, Lan Wang, and Yanmin Qian,
\newblock ``Data augmentation for end-to-end code-switching speech recognition,''
\newblock in {\em 2021 IEEE Spoken Language Technology Workshop (SLT)}.

\bibitem{ugan2022language}
Enes~Yavuz Ugan, Christian Huber, Juan Hussain, and Alexander Waibel,
\newblock ``Language-agnostic code-switching in sequence-to-sequence speech recognition,''
\newblock {\em arXiv preprint arXiv:2210.08992}, 2022.

\bibitem{nakayama2021code}
Sahoko Nakayama, Andros Tjandra, Sakriani Sakti, and Satoshi Nakamura,
\newblock ``Code-switching asr and tts using semisupervised learning with machine speech chain,''
\newblock {\em IEICE TRANSACTIONS on Information and Systems}, 2021.

\bibitem{pfeiffer2020adapterfusion}
Jonas Pfeiffer, Aishwarya Kamath, Andreas R{\"u}ckl{\'e}, Kyunghyun Cho, and Iryna Gurevych,
\newblock ``Adapterfusion: Non-destructive task composition for transfer learning,''
\newblock {\em arXiv preprint arXiv:2005.00247}.

\bibitem{pham2022adaptive}
Ngoc-Quan Pham, Alex Waibel, and Jan Niehues,
\newblock ``Adaptive multilingual speech recognition with pretrained models,''
\newblock {\em arXiv preprint arXiv:2205.12304}, 2022.

\bibitem{rolland2024shared}
Thomas Rolland and Alberto Abad,
\newblock ``Shared-adapters: A novel transformer-based parameter efficient transfer learning approach for children’s automatic speech recognition,''
\newblock in {\em Interspeech}, 2024.

\bibitem{liu2023kit}
Danni Liu, Thai~Binh Nguyen, Sai Koneru, Enes~Yavuz Ugan, et~al.,
\newblock ``Kit's multilingual speech translation system for iwslt 2023,''
\newblock {\em arXiv preprint arXiv:2306.05320}, 2023.

\bibitem{xuhong2018explicit}
LI~Xuhong, Yves Grandvalet, and Franck Davoine,
\newblock ``Explicit inductive bias for transfer learning with convolutional networks,''
\newblock in {\em ICML}, 2018.

\bibitem{kirkpatrick2017overcoming}
James Kirkpatrick, Razvan Pascanu, Neil Rabinowitz, et~al.,
\newblock ``Overcoming catastrophic forgetting in neural networks,''
\newblock {\em Proceedings of the national academy of sciences}, 2017.

\bibitem{zenke2017continual}
Friedemann Zenke, Ben Poole, and Surya Ganguli,
\newblock ``Continual learning through synaptic intelligence,''
\newblock in {\em ICML}, 2017.

\bibitem{poth2023adapters}
Clifton Poth, Hannah Sterz, Indraneil Paul, et~al.,
\newblock ``Adapters: A unified library for parameter-efficient and modular transfer learning,''
\newblock {\em arXiv preprint arXiv:2311.11077}, 2023.

\bibitem{fortunato2017bayesian}
Meire Fortunato, Charles Blundell, and Oriol Vinyals,
\newblock ``Bayesian recurrent neural networks,''
\newblock {\em arXiv preprint arXiv:1704.02798}, 2017.

\bibitem{blundell2015weight}
Charles Blundell, Julien Cornebise, Koray Kavukcuoglu, and Daan Wierstra,
\newblock ``Weight uncertainty in neural network,''
\newblock in {\em ICML}, 2015.

\bibitem{wang2024blob}
Yibin Wang, Haizhou Shi, Ligong Han, Dimitris Metaxas, and Hao Wang,
\newblock ``Blob: Bayesian low-rank adaptation by backpropagation for large language models,''
\newblock {\em Advances in Neural Information Processing Systems}, pp. 67758--67794.

\bibitem{yang2023bayesian}
Adam~X Yang, Maxime Robeyns, Xi~Wang, and Laurence Aitchison,
\newblock ``Bayesian low-rank adaptation for large language models,''
\newblock {\em arXiv preprint arXiv:2308.13111}, 2023.

\bibitem{xie2021bayesian}
Xurong Xie, Xunying Liu, Tan Lee, and Lan Wang,
\newblock ``Bayesian learning for deep neural network adaptation,''
\newblock {\em IEEE/ACM Transactions on Audio, Speech, and Language Processing}.

\bibitem{kingma2013auto}
Diederik~P Kingma and Max Welling,
\newblock ``Auto-encoding variational bayes,''
\newblock {\em arXiv preprint arXiv:1312.6114}, 2013.

\bibitem{ugan2025pier}
Enes~Yavuz Ugan, Ngoc-Quan Pham, Leonard B{\"a}rmann, and Alex Waibel,
\newblock ``Pier: A novel metric for evaluating what matters in code-switching,''
\newblock in {\em ICASSP}. IEEE, 2025.

\end{thebibliography}

\end{document}